\documentclass[lettersize,journal]{IEEEtran}

\usepackage{amsmath,amsfonts}
\usepackage{graphicx}
\usepackage{cite}
\usepackage{url}
\usepackage{textcomp}
\usepackage{stfloats}   
\usepackage{booktabs}   
\usepackage{multirow}
\usepackage{makecell}
\usepackage[table]{xcolor}
\definecolor{myblue}{RGB}{230,245,255}
\usepackage{cuted}      

\makeatletter
\def\@IEEEfigurecaptionsepspace{\vskip 0pt\relax}
\makeatother
\newcommand{\figtrim}[3]{\includegraphics[width=#1,trim=0 #2 0 0,clip]{#3}}

\usepackage{hyperref}
\hypersetup{colorlinks=true, citecolor=blue, linkcolor=blue, urlcolor=blue}

\begin{document}

\title{What Makes a Good Visual Representation for \\ Spatial Understanding in MLLMs?}

\author{Xiaohu Huang, Yumin Guo, and Kai Han%
\thanks{Xiaohu Huang, Yumin Guo, and Kai Han are with The University of Hong Kong. Corresponding author: Kai Han (e-mail: kaihanx@hku.hk).}}

\markboth{}%
{Huang \MakeLowercase{\textit{et al.}}: What Makes a Good Visual Representation for Spatial Understanding in MLLMs}

\maketitle

\begin{abstract}
Spatial understanding is a necessary step in moving multimodal large language models (MLLMs) toward reasoning about the 3D physical world. Their visual pretraining data, however, is almost entirely 2D, so the visual tokens processed by MLLMs often lack inherent 3D structural information. A growing number of methods inject such structure by transferring representations from external vision or geometry models, and collectively establish that such transfer helps. Behind these methods lies a largely underexplored question: what makes a visual representation good for spatial understanding? To answer it, we first separate what is transferred from how it is transferred, and then trace the resulting differences back to properties of the representations themselves. What we transfer comes from 3D geometry models, self-supervised image and video models, vision-language models, and generative models, in the form of latent features or explicit depth and point maps. We transfer it either by distillation, using it as a supervision target, or by fusion, feeding it as an additional input. We find that a representation's 3D structure correlates strongly with how much it helps, measured by cross-view consistency for latent features and by depth and reconstruction accuracy for explicit geometry.
We further show that the form of a representation dictates the mechanism it needs. Latent features gain as much from fusion as from distillation; explicit geometry gains little from fusion, and most of its distillation gain depends on a geometry decoding head. Guided by these findings, we distill both latent features and explicit geometry into a single MLLM. Taking only RGB video as input, it consistently improves over its base model on both static 3D and dynamic 4D spatial reasoning benchmarks.
Project page: \url{https://visual-ai.github.io/3drs}

\end{abstract}


\begin{strip}
\centering
\includegraphics[width=\textwidth]{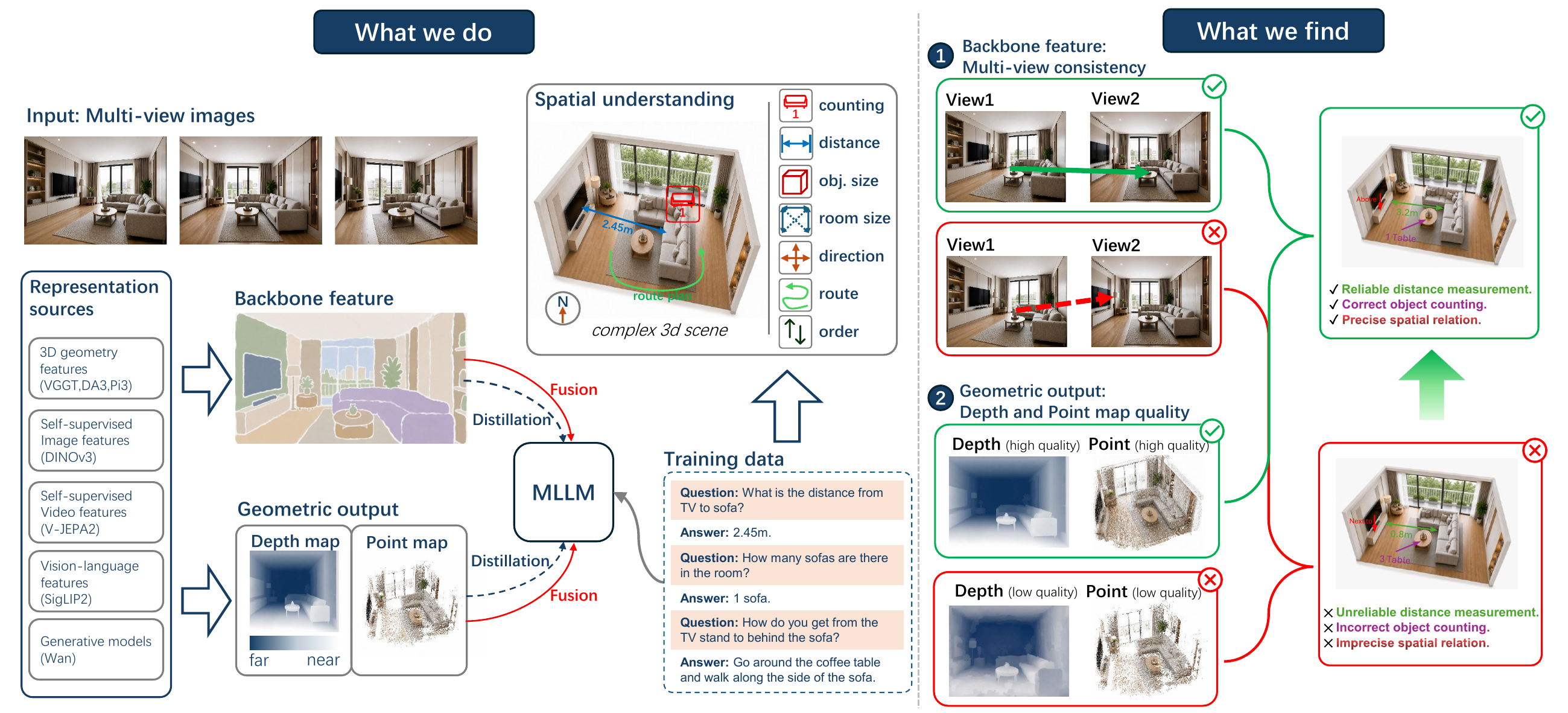}\\[5pt]
\refstepcounter{figure}\label{fig:teaser}%
\parbox{\textwidth}{\footnotesize Fig.~\thefigure.\quad \textbf{What makes a visual
representation good for spatial understanding?}
\emph{Left (what we do):} holding everything else fixed, we vary the representation
transferred into the MLLM---where it is drawn from, what form it takes, and how it enters
the model.
\emph{Right (what we find):} the gain a representation brings tracks the 3D structure it
carries, measured as cross-view consistency for backbone features and as depth and
reconstruction accuracy for geometric outputs.}
\end{strip}

\section{Introduction}

\IEEEPARstart{S}{patial} understanding aims to extend multimodal large language
models (MLLMs) from 2D perception to reasoning about 3D scenes. Modern MLLMs have made rapid progress by aligning
large language models with powerful image and video encoders~\cite{clip,flamingo,blip2,minigpt4,instructblip,llava,llava-onevision,llavavideo,qwen2vl,qwen25vl,internvl,gpt-4}.
These models are effective at recognizing visual content, but recognizing what is
present is not the same as understanding how a scene is structured in 3D space.

The difficulty lies in what the language model receives. Spatial reasoning requires
visual tokens that preserve 3D scene structure, not only object semantics. Recent
benchmarks consistently identify this as a persistent weakness of current
MLLMs~\cite{scanqa,sqa3d,video3dllm,cv_bench,spar,spatialscore,vsi_bench,viewspatial,dsr_suite}.
A root cause of this deficiency is the training supervision: most MLLMs are pretrained predominantly on 2D image-text pairs.
Consequently, their visual encoders target 2D semantic recognition, and the models receive little
direct 3D supervision. Spatial structure is often absent from the
tokens on which the language model operates. Improving spatial reasoning therefore
requires improving the visual representation itself.

A growing body of work seeks to supply this missing structure. Early methods build
the language interface on explicit 3D inputs, such as point clouds~\cite{pointllm,pointbind,gpt4point},
RGB-D scans~\cite{llava3d,video3dllm,gpt4scene}, object proposals~\cite{chatscene,inst3d,grounded3dllm},
or reconstructed scene representations~\cite{3d_llm,scenellm,ll3da}. These inputs
expose geometry directly, but they depend on 3D sensing or reconstruction that is
costly and often unavailable at inference. Recent work therefore favors MLLMs that
operate on plain image or video. Such methods incorporate spatial information through
auxiliary visual models or representations: spatial instruction data~\cite{spatialvlm,spatialrgpt,spatialbot,spar},
geometry-aware encoders~\cite{spatial_mllm,vg_llm}, reconstruction tokens~\cite{vlm3r},
progressive spatial training~\cite{spatialladder}, or geometry-language modules~\cite{spatialstack,guide,g2vlm}.
Despite their differences, these methods share a common move: they bring an external
visual representation, sometimes an explicit geometric one, into the MLLM.

Taken together, these methods have shown that an external representation can supply the
structure an MLLM lacks. However, the intrinsic properties of the representations themselves remain largely underexplored. The critical question, therefore, is not whether transfer is beneficial, but rather what intrinsically makes a visual representation effective for spatial understanding. Addressing this necessitates a rigorously controlled study.

In this paper, we conduct such a systematic investigation (Fig.~\ref{fig:teaser}, left).
Our study proceeds in two phases.
First, we decouple \emph{what} is transferred from \emph{how} it is transferred,
holding the underlying architecture and training recipe constant.
This ensures that any variance in spatial reasoning performance can be strictly attributed to these specific choices.
Second, we evaluate each representation offline---without downstream training---to identify which intrinsic metrics correlate with downstream performance gains.
The content of the transfer (``what'') comprises two dimensions.
The first is the source model:
candidates include visual geometry models~\cite{vggt,pi3,da3},
self-supervised image models~\cite{dinov3},
vision-language models~\cite{siglip2},
generative models~\cite{wan}, and
self-supervised video models~\cite{vjepa2},
each encoding 3D structure to varying degrees.
The second is the representation form: latent features extracted from model backbones,
versus explicit depth and point maps predicted exclusively by geometry models.
The transfer mechanism (``how'') involves a choice between \emph{distillation}
(utilizing the representation as a supervision target during training)
and \emph{fusion} (providing it as an auxiliary input during both training and inference).

On what to transfer, we find that a model's usefulness correlates strongly with the 3D
structure its representation carries~(Fig.~\ref{fig:teaser}, right). For latent features, we measure it
as cross-view consistency~\cite{probe3d,lexicon3d,multiview,3drs}; features that stay
consistent across viewpoints bring larger downstream gains, whereas 2D
feature-structure metrics correlate only weakly. For explicit geometry, we measure it as depth and
reconstruction accuracy; models that reconstruct a scene more accurately yield larger
gains once transferred. The two forms, in turn, specialize rather than
substitute: geometry contributes most on single-object metric questions such as
counting, object size, and absolute distance, while features contribute most on
multi-object relational and temporal ones. A candidate can therefore be assessed before
it is ever transferred.

On how to transfer it, we find that the form of a representation decides which mechanism
works: backbone
features benefit similarly from distillation and fusion, whereas explicit geometry is
transferred far better by distillation, especially when a Dense Prediction
Transformer (DPT)-style head from VGGT decodes the MLLM's own features into depth and
point maps.

Synthesizing these insights, we propose a unified design:
transferring both representation forms simultaneously via distillation,
utilizing the decoding head for geometry.
The resulting model requires no external auxiliary models during inference and achieves consistent improvements across all benchmarks compared to its base model.

A preliminary version of this work appeared as 3DRS~\cite{3drs} at NeurIPS 2025, where we showed
that the 3D-awareness of an MLLM's visual tokens correlates with scene understanding,
and that distilling features from a 3D foundation model strengthens it. This paper
extends that study in five aspects. First, we move from a single 3D
foundation model to representations drawn from five families of pretrained model.
Second, we cover both backbone features and explicit geometric outputs. Third, we compare
distillation and fusion, including the effect of the pretrained VGGT DPT-style
decoding head. Fourth, we relate representation traits---cross-view 3D correspondence
for features and depth and reconstruction accuracy for geometric outputs---to
downstream performance. Fifth, we broaden the evaluation with dynamic
(4D) spatial reasoning on DSR-Bench~\cite{dsr_suite} and comparisons with recent
spatial MLLMs.

The contributions of this paper are summarized as follows:
\begin{itemize}
    \item We cast spatial representation transfer for MLLMs as a controlled study.
    A transfer involves three decisions at once---which pretrained model supplies the
    spatial signal, what form that signal takes, and how it enters the MLLM. We separate
    them into three axes of a single framework, in which each can be varied while the
    others are held fixed, so that the effect of each becomes measurable on its own.

    \item Within this framework, we compare backbone-feature transfer against explicit
    geometric transfer, each through both distillation and fusion. Distillation is
    clearly the stronger mechanism for geometric representations, especially when paired
    with a pretrained geometry decoding head, whereas for backbone features the two
    mechanisms perform comparably. The two forms also specialize by spatial skill, which
    is why we transfer both rather than choose between them.

    \item We identify measurable properties that go with a visual representation being
    useful for spatial
    understanding. For backbone features, cross-view consistency correlates strongly with
    downstream spatial gains, while standard 2D feature-structure metrics correlate only
    weakly. For explicit geometry, the depth and reconstruction accuracy of the
    pretrained model tracks the gain it transfers.

    \item Guided by these empirical findings,
    we instantiate an optimal configuration that jointly distills latent features and explicit geometry into a single MLLM.
    This unified model leverages the complementary strengths of both forms without incurring additional inference costs:
    it strictly consumes RGB video, requires no auxiliary depth sensors or geometry models,
    and consistently outperforms its base model across both static 3D and dynamic 4D reasoning benchmarks.
\end{itemize}

\section{Related Work}
\label{sec:related}

\subsection{Spatial Understanding Benchmarks}
\label{sec:related_bench}

A growing set of benchmarks pushes the evaluation of MLLMs beyond object-level
semantics toward metric, relational, and viewpoint-dependent reasoning.
Early efforts ground language in 3D scenes through object localization and dense
captioning on ScanNet~\cite{scannet}, as in ScanRefer~\cite{scanrefer},
Multi3DRefer~\cite{multi3drefer}, and Scan2Cap~\cite{scan2cap}.
Question-answering suites built on indoor scans, such as ScanQA~\cite{scanqa} and
SQA3D~\cite{sqa3d}, probe situated reasoning over 3D scenes, while
VSI-Bench~\cite{vsi_bench} targets visual spatial intelligence from video.
Real-3DQA~\cite{real3dqa} further shows that SQA3D-style evaluation can contain
textual shortcuts, and proposes a filtered benchmark for testing 3D evidence use.
CV-Bench~\cite{cv_bench}, SPAR-Bench~\cite{spar}, and
ViewSpatial-Bench~\cite{viewspatial} isolate 2D/3D spatial relations and
cross-view localization, and SpatialScore~\cite{spatialscore} and
MMSI-Bench~\cite{mmsi_bench} aggregate diverse spatial skills into single
corpora. More recently, dynamic spatial reasoning has been formalized by
DSR-Bench~\cite{dsr_suite}, which evaluates reasoning over moving objects and
changing viewpoints. Across this suite, spatial reasoning remains among the least
reliable abilities of current MLLMs, motivating methods that make visual
tokens carry richer spatial structure. We adopt a representative subset of them
as a fixed evaluation suite so that different representation and transfer choices
can be compared under identical conditions.

\subsection{Scene Understanding with Explicit 3D Inputs}
\label{sec:related_3dinput}

One line of work equips language models with explicit 3D inputs. Early methods
align point-cloud and text features for 3D comprehension, including
PointLLM~\cite{pointllm}, Point-Bind~\cite{pointbind}, and
GPT4Point~\cite{gpt4point}. Building on this, 3D-LLM~\cite{3d_llm},
LL3DA~\cite{ll3da}, Grounded~3D-LLM~\cite{grounded3dllm},
Inst3D-LLM~\cite{inst3d}, and Scene-LLM~\cite{scenellm} design cross-modal
interfaces over reconstructed scenes, object proposals, or scene graphs, while
Chat-Scene~\cite{chatscene} exploits complementary 2D and 3D features.
A related thread pre-trains aligned 3D vision-language representations or builds
embodied agents over 3D scenes, including 3D-VisTA~\cite{3dvista},
SceneVerse~\cite{sceneverse}, and the embodied generalist LEO~\cite{leo}.
MLLM-style pipelines such as LLaVA-3D~\cite{llava3d}, Video-3D
LLM~\cite{video3dllm}, and GPT4Scene~\cite{gpt4scene} inherit the image/video
backbone but still rely on 3D priors derived from depth, camera pose, point
clouds, or voxelized scans. These methods establish that explicit geometry is
highly informative for scene understanding; however, they assume access to
external 3D or 2.5D inputs at inference, placing them in a different input
regime from methods that operate purely from RGB video. We therefore treat them
as a separate setting rather than as direct comparisons, and report the input
protocol explicitly to keep comparisons fair.

\subsection{Spatial MLLMs with Internal Geometry Priors}
\label{sec:related_geoprior}

A second line keeps the input close to RGB images or video but strengthens the
model with internally estimated geometry. Spatial-MLLM~\cite{spatial_mllm}
couples a semantic encoder with a geometry encoder initialized from a visual
geometry model, and VG-LLM~\cite{vg_llm} integrates 3D vision-geometry priors
learned from monocular video. VLM-3R~\cite{vlm3r} augments the language model
with instruction-aligned reconstruction tokens, while
SpatialStack~\cite{spatialstack}, GUIDE~\cite{guide}, and
G$^2$VLM~\cite{g2vlm} explore layered or layer-wise geometry-language fusion and
unified reconstruction-and-reasoning objectives. Although these methods differ in
architecture and supervision, they share a common ingredient: external visual or
geometric representations are brought into the MLLM, as additional tokens, fused
features, or supervision targets. Each, however, is evaluated within its own
architecture and training recipe, so the contribution of the representation itself is
left open. Our study is
complementary: rather than proposing one more recipe, we hold the base model and
protocol fixed and vary these factors one at a time.

\subsection{Spatial Instruction Data and Training}
\label{sec:related_training}

Spatial ability can also be improved through training data and recipes.
SpatialVLM~\cite{spatialvlm}, SpatialRGPT~\cite{spatialrgpt}, and
SpatialBot~\cite{spatialbot} construct large-scale spatial instruction data for
supervised fine-tuning, and SpatialLadder~\cite{spatialladder} adopts progressive
spatial training. Like these methods, our study stays within a standard
supervised fine-tuning regime, and is orthogonal to lines that instead rely on
reinforcement learning or inference-time reasoning---such as SpaceR~\cite{spacer},
ViLaSR~\cite{vilasr}, Video-R1~\cite{video-r1}, and 3DThinker~\cite{3dthinker}---or
that target robotic affordance prediction~\cite{robopoint}. Our contribution is
complementary to all of these: under a fixed training recipe, we ask which
pretrained representation should be transferred, and how.

\subsection{3D-Aware Representation Transfer}
\label{sec:related_transfer}

Our study builds on visual foundation models and analyses of their 3D-awareness.
Image-level foundation models---CLIP~\cite{clip}, DINO~\cite{dino},
MAE~\cite{mae}, DINOv2~\cite{dinov2}, DINOv3~\cite{dinov3}, SigLIP2~\cite{siglip2},
and SAM~\cite{sam}---provide strong but largely 2D-semantic features, while
self-supervised video models such as V-JEPA2~\cite{vjepa2} and generative models
such as Wan~\cite{wan} supply representations from two further pretraining
paradigms. In parallel, visual
geometry models predict
explicit geometry from images or video, ranging from monocular depth and geometry
estimators such as MiDaS~\cite{midas}, Metric3D~\cite{metric3d},
UniDepth~\cite{unidepth}, MoGe~\cite{moge}, and Depth
Anything~\cite{depth_anything,depthanythingv2}, to multi-view reconstruction
models such as DUSt3R~\cite{dust3r}, MASt3R~\cite{mast3r},
Spann3R~\cite{spann3r}, Fast3R~\cite{fast3r}, CUT3R~\cite{cut3r},
VGGT~\cite{vggt}, FLARE~\cite{flare}, $\pi^3$~\cite{pi3}, and DA3~\cite{da3}.
Probe3D~\cite{probe3d}, Lexicon3D~\cite{lexicon3d}, and multi-view
analyses~\cite{multiview} measure how much 3D structure such representations
encode, mostly for pure vision models. Transferring knowledge across models by
distillation~\cite{distill} is a classic tool; the open question we study is
which of these representations are worth transferring for spatial understanding,
and how. Our conference paper 3DRS~\cite{3drs} extended this view to MLLMs,
supervising their visual tokens with a pretrained 3D foundation model and linking the
resulting 3D-awareness to downstream performance. This journal version generalizes that single
supervision setting into a controlled comparison across representation families,
representation forms (backbone features vs.\ explicit geometric outputs), and
transfer mechanisms (distillation vs.\ fusion), and it analyzes which measurable
representation traits track the gains.

\section{Spatial Representation Learning: From Analysis to Design}
\label{sec:method}

\begin{figure*}[!t]
\centering
\figtrim{0.92\textwidth}{6.74}{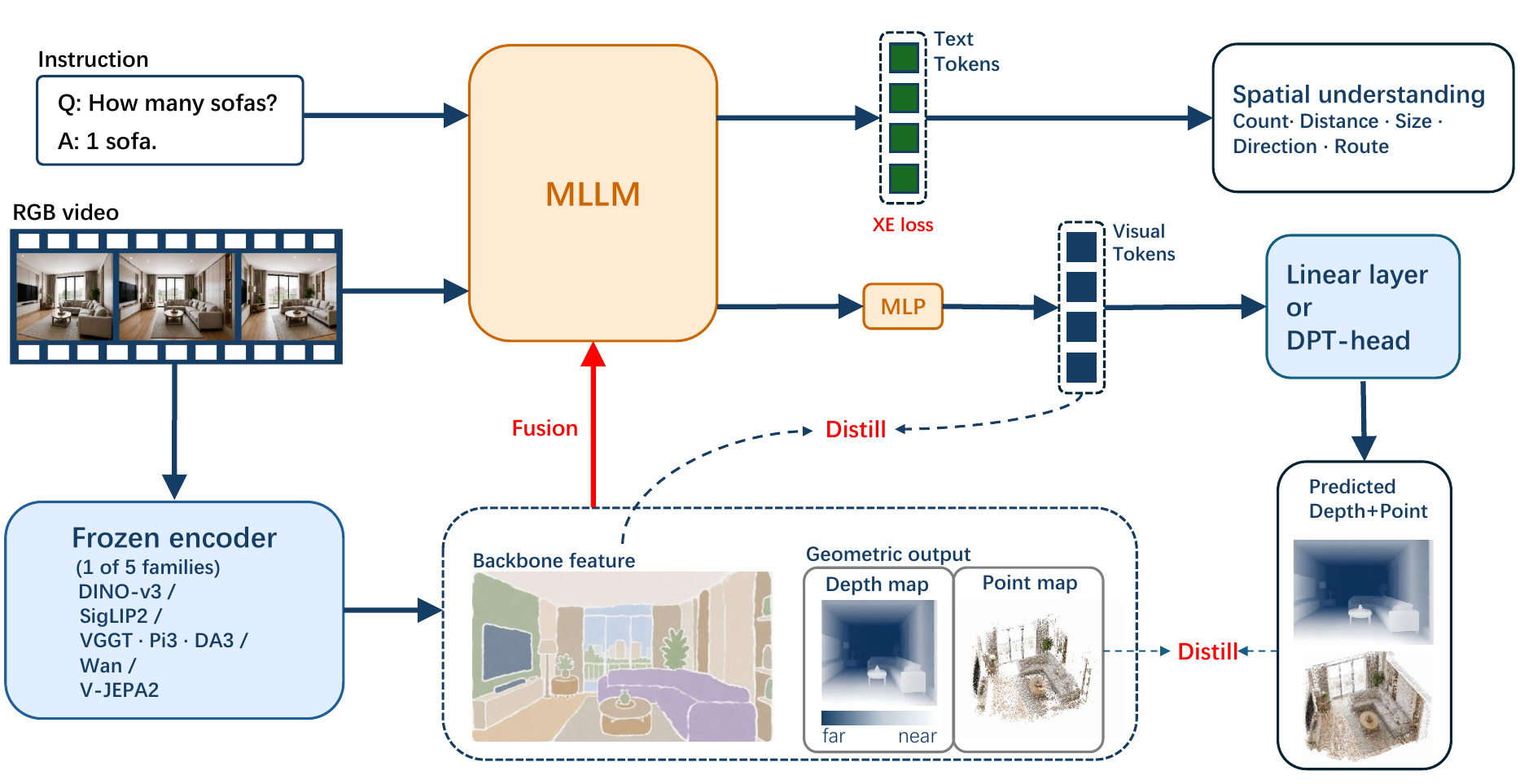}
\caption{\textbf{Overview of the study.} We transfer a
representation from a pretrained model---one of five families---into an
MLLM whose base model is fixed across all variants. The
representation takes one of two forms, backbone features or explicit
geometric outputs, and enters the MLLM by one of two mechanisms,
distillation or fusion. The MLLM
itself takes only RGB video and is evaluated on spatial understanding.}
\label{fig:framework}
\end{figure*}

An MLLM's spatial reasoning capability is fundamentally constrained by the degree of 3D
structure encoded within its visual tokens. While external representations can compensate
for this structural deficit, the extent of the enhancement varies significantly across
different representations. This section investigates the underlying drivers of this
variation.

Specifically, this variance stems from three design choices that fully define a transfer:
the \emph{source} pretrained model, the representation \emph{form} (i.e., latent
\emph{backbone features} versus explicit \emph{geometric outputs} such as depth or point
maps), and the integration \emph{mechanism} (distillation versus fusion). By systematically
varying each dimension while holding the others constant~(Fig.~\ref{fig:framework}), we
identify the intrinsic properties of a representation that dictate its downstream
performance.

The remainder of this section is structured as follows.
Section~\ref{sec:method_diagnostics} addresses the selection of the pretrained model
through offline measurements conducted on the representations prior to any downstream
training. Section~\ref{sec:method_typespec} delineates the distinct spatial contributions
of the two representation forms. Subsequently, Section~\ref{sec:method_transfer} evaluates
the transfer mechanisms, revealing that the optimal choice is inherently constrained by the
representation's form. Finally, Section~\ref{sec:method_design} synthesizes these empirical
insights into a unified, optimal architectural design.

\subsection{Setup and Protocol}
\label{sec:method_protocol}

An MLLM consists of a visual encoder $\mathcal{E}_\mathrm{img}$ and a language
decoder $\mathcal{T}$. Given a video or multi-view image sequence
$\mathcal{I}=\{I_1,\dots,I_N\}$ and a language instruction, the encoder produces
per-frame visual features $\mathbf{F}_i\in\mathbb{R}^{H\times W\times d}$ that the
decoder consumes together with the instruction tokens to generate the answer. We denote
an external pretrained model by $g$, which maps the same input $\mathcal{I}$ to a
representation $\mathbf{R}=g(\mathcal{I})$.

\noindent\textbf{Fixed protocol.}
We keep the base MLLM, the optimizer and schedule, the visual-token budget, the
training data, and the evaluation suite identical across all variants; each
experiment changes one of the three choices at a time, so that any difference in
downstream performance can be attributed to that choice. Training stays within a
standard supervised fine-tuning (SFT) regime: the MLLM is optimized with the
autoregressive cross-entropy loss $\mathcal{L}_\mathrm{CE}$, to which a
transfer-specific term is added as defined below. We deliberately avoid
reinforcement learning and inference-time reasoning, so that the comparison
reflects the transferred representation rather than the training paradigm.
Unless stated otherwise, representations are transferred by distillation; the choice of
mechanism is itself examined in Section~\ref{sec:method_transfer}.

\noindent\textbf{Pure-2D input.}
The MLLM receives \emph{only} RGB video. Any geometry is either
internalized through distillation (and thus unnecessary at inference) or produced
on the fly by a visual geometry model from the same RGB frames. Depth and pose,
when used, enter only the \emph{offline} diagnostics of
Section~\ref{sec:method_diagnostics}, never the model input.

\noindent\textbf{Decoding geometric outputs.}
Backbone features can be matched to the MLLM's tokens directly, but depth and point maps
cannot: they must be decoded from those tokens. We use a pretrained DPT-style decoding
head from VGGT~\cite{vggt}. A DPT decoder consumes \emph{multi-layer} features rather
than a single layer, so we feed it four evenly spaced MLLM layers ($\{2,10,18,31\}$) as
its input. We keep the head at its pretrained weights, so the MLLM is trained to produce
features that this geometry decoder can turn into depth and point maps, leaving
fine-grained geometry to the decoder. The same head serves every geometry model we
transfer from: it decodes the MLLM's own features rather than the teacher's, so it is
not tied to the model supplying the target. As an alternative we
replace it with a \emph{learnable} lightweight linear head, which must instead force the
MLLM features themselves to carry metric depth and 3D position.

\noindent\textbf{Reporting.}
Throughout this section a gain is reported as relative improvement over the base model,
$(\text{variant}-\text{base})/\text{base}$, averaged over the evaluation suite; the
comparison tables of Section~\ref{sec:experiments} report absolute scores instead.

\subsection{3D Structure Tracks What a Model Contributes}
\label{sec:method_diagnostics}

We draw pretrained models $g$ from five families that span the
pretraining paradigms used by recent spatial MLLMs:
(1)~\textbf{visual geometry models}---VGGT~\cite{vggt}, $\pi^3$~\cite{pi3}, and
DA3~\cite{da3};
(2)~\textbf{self-supervised image models}---DINOv3~\cite{dinov3};
(3)~\textbf{vision-language models}---SigLIP2~\cite{siglip2};
(4)~\textbf{generative models}---Wan~\cite{wan}; and
(5)~\textbf{self-supervised video models}---V-JEPA2~\cite{vjepa2}.
Their gains differ widely. To find what those differences track, we measure intrinsic
traits of each representation \emph{offline}---never feeding them to the MLLM---and relate
those measurements to the downstream gain, separately for the two forms.

\begin{figure*}[!t]
\centering
\figtrim{0.82\textwidth}{3.23}{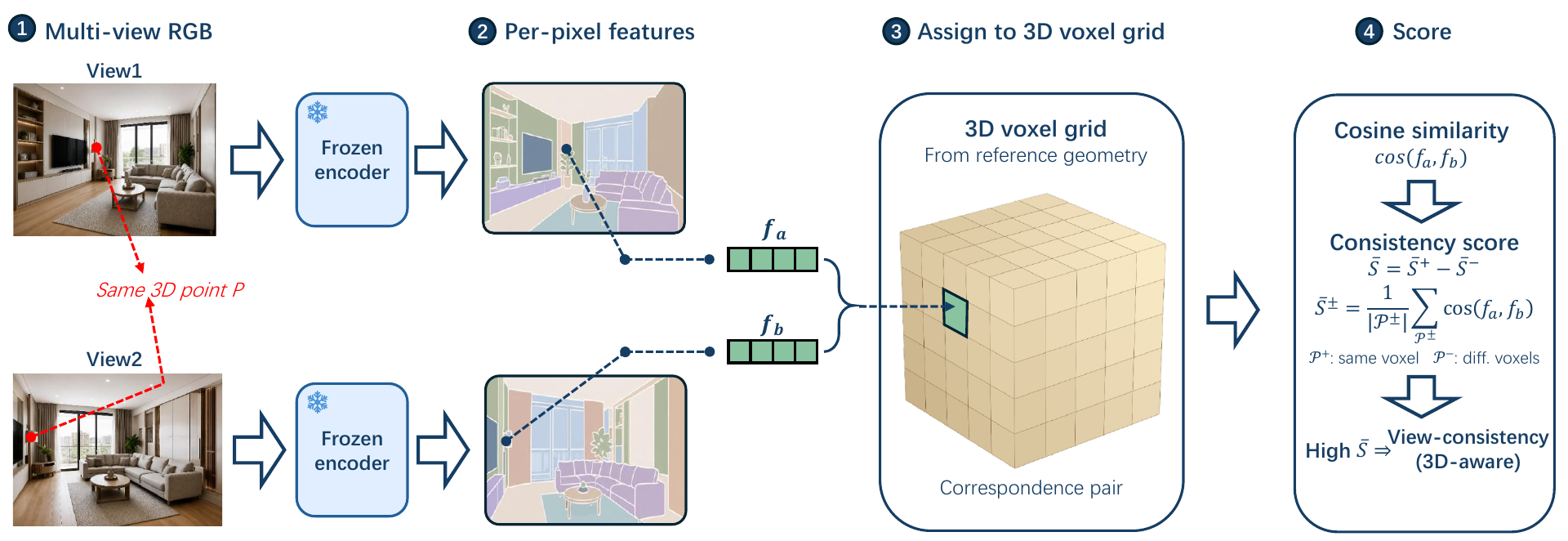}
\caption{\textbf{Measuring 3D-aware cross-view consistency.} Per-pixel features from
different views are binned into a shared 3D voxel grid. The score $\bar S$ is how far
genuine correspondences stand above mismatched pairs in mean cosine similarity, so a
representation that fails to distinguish locations scores near zero.}
\label{fig:crossview}
\end{figure*}

\noindent\textbf{Backbone features: 3D-aware cross-view consistency.}
For a \emph{backbone feature}, we measure 3D structure through cross-view correspondence,
a fundamental property of 3D geometry~\cite{mvg} whose consistency can be computed
directly from a representation's features, without training any probe.
Concretely, we voxelize each scene using the sensor depth and camera poses that ship with
the source scans, and assign every per-pixel feature to a voxel by its 3D location
(Fig.~\ref{fig:crossview}). Features from different views that fall into the same voxel
are genuine correspondences $\mathcal{P}^{+}$; features drawn from different voxels are
mismatches $\mathcal{P}^{-}$. Mean same-voxel similarity alone would not measure 3D
structure: a representation that returned one constant vector everywhere would attain the
maximum without distinguishing any two locations. We therefore score a representation by
how far genuine correspondences stand \emph{above} mismatched ones,
\begin{equation}
  \bar{S}
  = \underbrace{\frac{1}{|\mathcal{P}^{+}|}\!\!\sum_{(\mathbf{f}_a,\mathbf{f}_b)\in\mathcal{P}^{+}}\!\!\!
      S(\mathbf{f}_a,\mathbf{f}_b)}_{\bar{S}^{+}}
  \;-\;
  \underbrace{\frac{1}{|\mathcal{P}^{-}|}\!\!\sum_{(\mathbf{f}_a,\mathbf{f}_b)\in\mathcal{P}^{-}}\!\!\!
      S(\mathbf{f}_a,\mathbf{f}_b)}_{\bar{S}^{-}},
\end{equation}
evaluated at several voxel resolutions. A higher $\bar{S}$ means the same 3D point is
represented consistently as the camera moves \emph{and} distinguishably from other points.
Wan, for instance, reaches $\bar{S}^{+}\!=\!0.90$ yet scores $\bar{S}\!=\!0.03$, being
nearly as similar across different 3D points as across the same one. As a further control, we compute three 2D feature-structure metrics~\cite{repaspatial}
that ignore 3D correspondence---local-distance, cross-distance, and relative-magnitude
structure consistency (LDS/CDS/RMSC)---which we return to below.

\begin{figure*}[!t]
\centering
\figtrim{0.8\textwidth}{6.58}{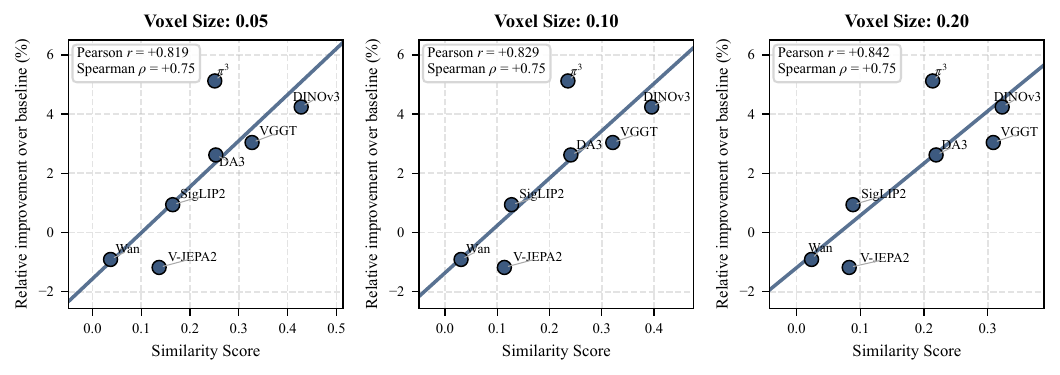}
\caption{\textbf{Backbone features: 3D-aware consistency tracks the gain.} Across
models, the cross-view consistency score correlates strongly with the gain from
distilling the feature (Pearson $r\!\approx\!0.83$), and the trend holds across voxel
resolutions.}
\label{fig:feat_3d}
\end{figure*}

\noindent\textbf{Finding~1:} \emph{3D-aware consistency tracks feature gains.}
Across the seven models, this cross-view consistency score is strongly correlated
with the spatial gain obtained by distilling the corresponding feature
(Fig.~\ref{fig:feat_3d}, Pearson $r\!\approx\!0.83$, Spearman
$\rho\!=\!0.75$), and the relationship is stable across voxel resolutions. No single
model carries it: removing any one of the seven leaves $r$ between $0.74$ and $0.95$. What tracks usefulness is therefore cross-view 3D consistency,
a property of the representation itself. Such consistency is not the preserve of geometry
models: DINOv3~\cite{dinov3}, a leading large-scale self-supervised image model, learns
highly discriminative features that are \emph{also} strongly 3D-consistent, and it
delivers one of the largest feature-transfer gains among the non-geometry models.
Large-scale self-supervised pretraining can therefore yield features that serve spatial
understanding well, and the diagnostic picks them out without reference to how they were
trained.

\begin{figure*}[!t]
\centering
\figtrim{0.8\textwidth}{8.26}{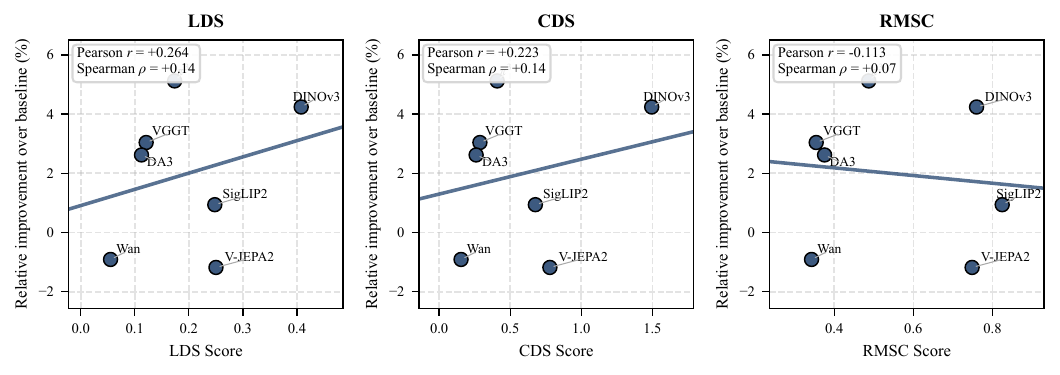}
\caption{\textbf{2D feature-structure metrics do \emph{not} track the gain.} Unlike
3D-aware cross-view consistency (Fig.~\ref{fig:feat_3d}), the three 2D metrics (LDS/CDS/RMSC)
are at most weakly correlated with the gain ($r\!\approx\!0.26,\,0.22,\,{-}0.11$).}
\label{fig:feat_2d}
\end{figure*}

The 2D control introduced above shows the correlation to be specific to 3D structure
(Fig.~\ref{fig:feat_2d}). The three 2D feature-structure metrics (LDS/CDS/RMSC), which
capture how regular a feature map is within each view but ignore cross-view 3D
correspondence, are at most weakly correlated with the downstream gain
($r\!\approx\!0.26,\,0.22,\,{-}0.11$). The contrast with the cross-view result
(Fig.~\ref{fig:feat_3d}) is the crux of the feature analysis: the gains track 3D-aware
structure, not generic feature regularity.

\noindent\textbf{Geometric outputs: depth and reconstruction quality.}
The same question has a different measurement for \emph{explicit geometry}. Here the
counterpart of cross-view consistency is how faithfully a model actually recovers 3D
structure: one whose predicted depth and point maps are more accurate carries more
reliable 3D information to transfer. We quantify this offline---again without feeding any
geometry to the MLLM---by scoring each model's native predictions on standard benchmarks
with two depth metrics (AbsRel on 7Scenes and an indoor average) and two reconstruction
metrics (F-Score on ScanNet++, F1 on ETH3D), and relate each score to the spatial gain
from distilling that model.

\begin{figure*}[!t]
\centering
\figtrim{0.85\textwidth}{7.34}{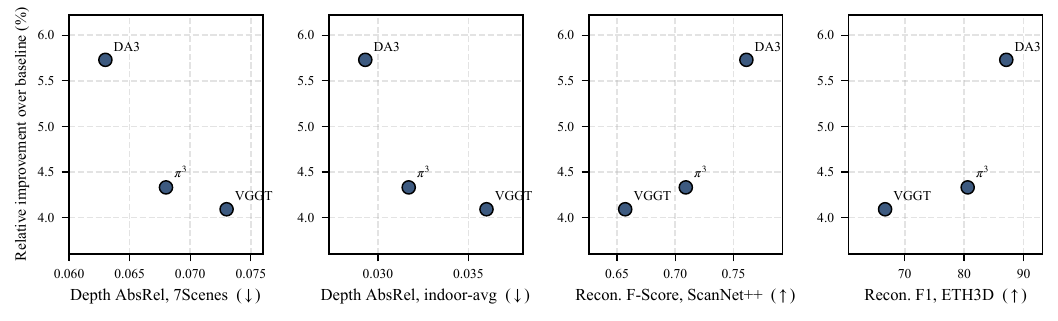}
\caption{\textbf{Explicit geometry: depth and reconstruction quality track the gain.}
Each panel scores the three geometry models' own predictions on a standard depth or
reconstruction benchmark against the gain from distilling them. The three fall in the
same order on all four.}
\label{fig:geo_quality}
\end{figure*}

\noindent\textbf{Finding~2:} \emph{Reconstruction quality tracks geometry gains.}
The three geometry models rank identically on all four depth and reconstruction metrics,
and in that same order by downstream gain (Fig.~\ref{fig:geo_quality}). With three models
the ordering is what the comparison supports, and it is the recurrence of one ordering
over four independent metrics---rather than any single fit---that carries the point. The
metrics are dense depth and reconstruction ones, matched to the kind of output the DPT
head has to produce. Reconstruction accuracy also organises performance inside the
transferred model itself: the scenes whose geometry it decodes most accurately are the
ones it answers most accurately, with QA accuracy falling from $64.0\%$ on the
best-reconstructed quarter of the evaluation scenes to $53.9\%$ on the worst
(Fig.~\ref{fig:geo_mech}c).

\emph{Although they measure different quantities, the two diagnostics play the same role.}
Cross-view correspondence assesses backbone features, while depth and reconstruction
accuracy gauge geometric outputs. Both confirm that representations with stronger inherent
3D structure yield larger downstream benefits.

\subsection{The Two Forms Overlap but Specialize}
\label{sec:method_typespec}

Knowing that what matters is the 3D structure a representation carries, we next ask in
which form that structure is better delivered. A backbone feature and a geometric output
drawn from the same model describe the same scene, so a natural question is whether they
strengthen the \emph{same} spatial skills or different ones. Holding the mechanism fixed
to distillation, we compare a backbone feature against an explicit geometric
output decoded by the DPT head, broken down by VSI-Bench task category and averaged over
the three geometry models (VGGT/DA3/$\pi^3$). The breakdown uses the ten task types the
benchmark scores separately; Table~\ref{tab:vsi} reports eight categories, the difference
being that the three difficulty splits of relative direction are averaged into one there.
We group the ten as \emph{single-object metric} (counting, object size, room size,
absolute distance), \emph{multi-object relational} (relative distance, and the three
relative-direction splits), \emph{temporal sequence} (appearance order), and
\emph{long-horizon} (route planning).

\begin{figure*}[!t]
\centering
\figtrim{0.72\textwidth}{6.60}{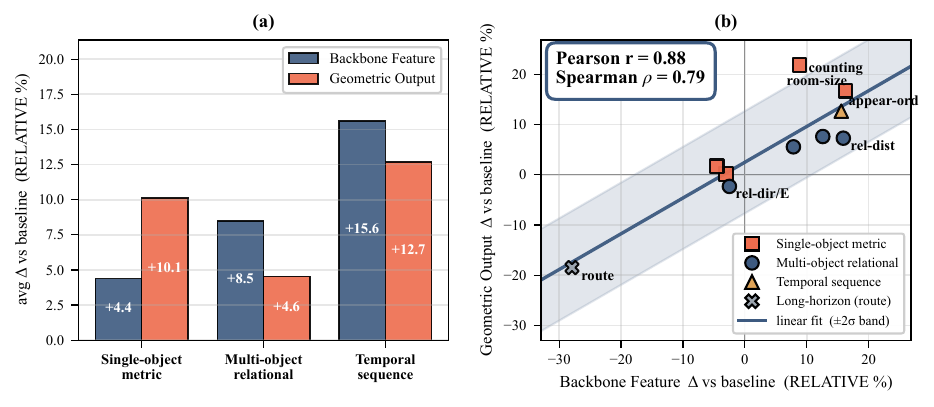}
\caption{\textbf{Backbone features and geometric outputs specialize by task.} Using
distillation, we compare backbone features against geometric outputs across tasks,
averaged over three geometry models (VGGT/DA3/$\pi^3$).
\emph{(a)}~Mean relative improvement over the base model by VSI-Bench task category.
\emph{(b)}~The two forms' per-task gains are strongly correlated (Pearson $r=0.88$):
single-object tasks favor geometric outputs (above the diagonal), relational/temporal
tasks favor backbone features (below).}
\label{fig:typespec}
\end{figure*}

\noindent\textbf{Finding~3:} \emph{Backbone features and geometric outputs specialize
by spatial skill.}
Across the ten task categories (Fig.~\ref{fig:typespec}), the two forms are strongly
correlated (Pearson $r=0.88$, Spearman $\rho=0.79$): they help and hurt largely the same
tasks. A shared task profile is consistent with the two forms carrying much the same
geometry, though it does not establish it; what the correlation rules out is that they
act as independent cues. What separates them is the systematic deviation from that trend. Geometric outputs, which carry
pixel-accurate decoded depth and point maps, give the larger gain on
\emph{single-object metric} tasks (counting, object and room size, absolute distance:
$+10.1\%$ vs.\ $+4.4\%$). Backbone features, being free-form and composable by the
language model, instead win \emph{multi-object relational} reasoning (relative distance
and direction: $+8.5\%$ vs.\ $+4.6\%$) and \emph{temporal-sequence} reasoning
(appearance order: $+15.6\%$ vs.\ $+12.7\%$).

\emph{The form decides not only how much a representation helps, but which spatial
ability it strengthens.} Neither form covers what the other does best, so we transfer
both.

\subsection{The Form Dictates the Mechanism}
\label{sec:method_transfer}

If both forms are worth transferring, the remaining choice is how each should enter the
MLLM. We compare the two mechanisms defined above, applying each to both forms.

\begin{figure}[!tb]
\centering
\figtrim{\columnwidth}{8.39}{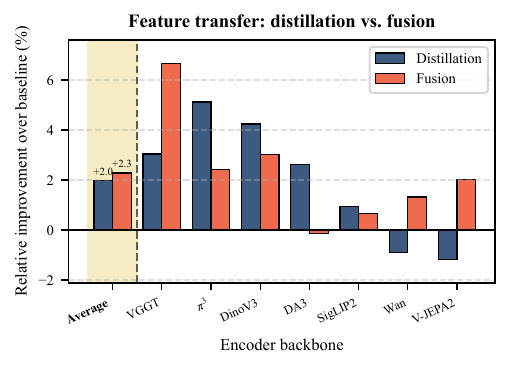}
\caption{\textbf{Backbone-feature transfer: distillation vs.\ fusion.} Bars are relative
improvement over the base model, per model. The two mechanisms yield broadly similar
gains for backbone features, with no consistent winner.}
\label{fig:feat_mech}
\end{figure}

\begin{figure*}[!t]
\centering
\figtrim{0.8\textwidth}{6.58}{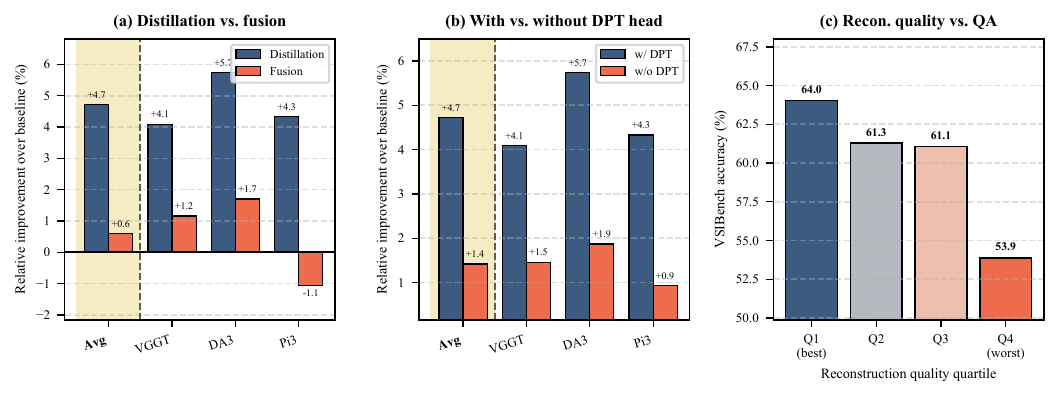}
\caption{\textbf{Explicit-geometry transfer and its quality.}
\emph{(a)}~Distillation far outperforms fusion for geometric outputs ($\approx$+4.7\%
vs.\ +0.6\%).
\emph{(b)}~Without the pretrained DPT-style decoding head, the gain falls to
$\approx$+1.4\%.
\emph{(c)}~Sorting the $288$ VSI-Bench evaluation scenes ($150$ from ARKitScenes, $88$
from ScanNet, $50$ from ScanNet++) by depth+point reconstruction error into quartiles,
QA accuracy falls from Q1 (64.0\%) to Q4 (53.9\%).}
\label{fig:geo_mech}
\end{figure*}

\noindent\textbf{Distillation.}
Following 3DRS~\cite{3drs}, we align the MLLM's per-pixel visual features with the
transferred representation. Each MLLM feature $\mathbf{f}_i(p)$ is projected by a
lightweight head $\mathrm{MLP}_\mathrm{align}$ and matched to the target
$\mathbf{r}_i(p)$ with a Smooth L1 regression objective,
\begin{equation}
  \mathcal{L}_\mathrm{distill}
  = \frac{1}{NHW}\sum_{i=1}^{N}\sum_{p\in I_i}
    \ell_{\mathrm{SL1}}
    \!\left(\mathrm{MLP}_\mathrm{align}(\mathbf{f}_i(p)),\,\mathbf{r}_i(p)\right),
\end{equation}
where $\ell_{\mathrm{SL1}}$ denotes the Smooth L1 loss, applied element-wise and
averaged over feature channels. For geometric outputs, $\mathbf{r}_i(p)$ is the depth or
point-map value predicted by the geometry model and the MLLM side is the corresponding
output of the DPT head. The targets are precomputed offline, so distillation adds no
inference-time cost and needs no external representation at test
time; the total objective is
$\mathcal{L}=\mathcal{L}_\mathrm{CE}+\mathcal{L}_\mathrm{distill}$.

\noindent\textbf{Fusion.}
Under fusion, the representation is supplied to the MLLM at both
training and inference through a learned \emph{gating} interface. Let
$\mathbf{v}_i(p)$ be the vision-encoder token at pixel $p$ of frame $i$ and let
$\mathbf{r}_i(p)$ be the external representation projected into the
token space by a linear map $W$. We form a gate from \emph{both} streams and add the
gated external representation back to the vision token as a residual,
\begin{equation}
  \mathbf{a}_i(p)=\sigma\!\big(\mathrm{MLP}\big([\mathbf{v}_i(p);\,W\mathbf{r}_i(p)]\big)\big),
\end{equation}
\begin{equation}
  \tilde{\mathbf{v}}_i(p)=\mathbf{v}_i(p)+\mathbf{a}_i(p)\odot W\mathbf{r}_i(p),
\end{equation}
where $[\cdot\,;\cdot]$ denotes concatenation, $\sigma$ the logistic function, and
$\odot$ the elementwise product. The gated tokens $\tilde{\mathbf{v}}$ replace
$\mathbf{v}$ as input to the decoder, so the model learns---per token and per
channel---how much of the external representation to admit. Unlike distillation, fusion
keeps the external representation available at test time, but it requires the external model
(or its geometry head) to run at inference.

\noindent\textbf{Finding~4:} \emph{For backbone features, the two mechanisms are
broadly comparable.}
For backbone features (Fig.~\ref{fig:feat_mech}), distillation and fusion give
comparable spatial gains once averaged over the evaluation suite, and \emph{which} of the
two wins is a property of the \emph{model} rather than of the mechanism: fusion is
slightly ahead for some backbones (e.g.,~VGGT) and distillation for others
(e.g.,~$\pi^3$), with no systematic gap between the two columns. The practical consequence is
that distillation---which needs no external model at inference---is the cheaper default
for backbone features, and fusion is worth its added test-time cost only when the external
model is already available at inference for other reasons.

\noindent\textbf{Finding~5:} \emph{Explicit geometry is best transferred by
distillation through a decoding head.}
For explicit geometric outputs, the two mechanisms behave very differently
(Fig.~\ref{fig:geo_mech}). Distillation lifts spatial accuracy by $\approx$+4.7\% on
average, whereas fusing the \emph{same} geometry adds only $\approx$+0.6\% and can
even hurt (panel~a): handing raw depth and point maps to the language model as an input
stream forces it to parse a representation it is poorly equipped to use,
while distillation lets it absorb that geometry into its own tokens during training
and then discard the external model. The benefit further hinges on the pretrained VGGT
DPT-style decoding head: replacing it with the learnable linear head reduces the
distillation gain to $\approx$+1.4\% (panel~b). A linear prediction head places the
burden of low-level, metric regression---precise depth values and 3D point
coordinates---inside the MLLM features themselves. The pretrained DPT head instead performs
this fine-grained geometry decoding from multi-layer MLLM features, so the MLLM can
supply higher-level visual representations without directly solving dense geometric
regression. The two heads differ in architecture, initialization and trainability at
once, so this comparison shows that the pretrained decoder configuration matters without
isolating which of the three is responsible.

\emph{The mechanism is therefore not a free choice.} Backbone features can be distilled
or fused with similar effect, so we distill them, which costs nothing at inference. For
geometric outputs the two routes are far apart: among the interfaces we implement,
distillation through the pretrained decoding head is the only one that delivers most of
the available gain.

\subsection{The Resulting Model}
\label{sec:method_design}

Across the three analyses one picture emerges. What a transfer contributes is the 3D
structure the representation carries; the two forms carry that structure in overlapping
but not interchangeable ways; and the mechanism that suits a form is not the same for
both. An effective design therefore has to do three things at once: draw on
representations rich in 3D structure, use both forms rather than one, and route each by
the mechanism that works for it.

Our model does exactly this. Both transferred representations are taken from
VGGT~\cite{vggt}: a backbone feature and an explicit geometric output
(Section~\ref{sec:method_typespec}), transferred by distillation, geometry through the
pretrained DPT-style head (Section~\ref{sec:method_transfer}), so that no external model
is needed at inference.

Concretely, the objective adds two transfer terms to the autoregressive loss,
\begin{equation}
  \mathcal{L}=\mathcal{L}_\mathrm{CE}
  +\mathcal{L}^\mathrm{feat}_\mathrm{distill}
  +\mathcal{L}^\mathrm{geo}_\mathrm{distill},
\end{equation}
where $\mathcal{L}^\mathrm{feat}_\mathrm{distill}$ aligns the MLLM's visual tokens with a
backbone feature and $\mathcal{L}^\mathrm{geo}_\mathrm{distill}$ aligns the depth and
point maps decoded from those tokens with a geometry model's predictions. The targets are
precomputed offline, so the resulting model consumes only RGB video at inference. This is
the configuration reported as \emph{Ours} throughout Section~\ref{sec:experiments}.

\section{Experiments}
\label{sec:experiments}

\begin{table*}[!tb]
\begin{minipage}[t]{0.44\textwidth}
\centering
\footnotesize
\setlength{\tabcolsep}{5pt}
\renewcommand{\arraystretch}{1.05}
\caption{Shared training configuration for the controlled comparisons; across variants,
only the transferred representation or transfer mechanism changes.}
\label{tab:impl}
\begin{tabular}{ll}
\toprule
Setting & Value \\
\midrule
Base MLLM           & Qwen3-VL-8B-Instruct \\
Framework           & LLaMA-Factory \\
Optimizer           & AdamW ($\beta_1{=}0.9,\beta_2{=}0.999$) \\
Adam $\epsilon$     & $1\times10^{-8}$ \\
Weight decay        & $0$ \\
Gradient clipping   & $1.0$ \\
Peak learning rate  & $1\times10^{-5}$ \\
LR schedule         & cosine ($\to 0$) \\
Warmup ratio        & $0.02$ \\
Epochs              & $1$ \\
GPUs                & $64$ \\
Per-device batch    & $1$ \\
Total batch size    & $64$ \\
Precision           & bf16 \\
Transfer tap layers & $\{2,10,18,31\}$ \\
Adapter width       & $2048$ \\
\bottomrule
\end{tabular}
\end{minipage}\hfill
\begin{minipage}[t]{0.54\textwidth}
\centering
\footnotesize
\setlength{\tabcolsep}{4pt}
\renewcommand{\arraystretch}{1.1}
\caption{Component ablation under the shared training recipe. ``Feat. distill'' distills a
backbone feature; ``Geo. distill'' distills a geometric output through the DPT head;
``Both'' combines the two and is the model reported as \emph{Ours}. All values are accuracy
(\%) except ScanQA (CIDEr); Avg.\ excludes ScanQA. Best per row in \textbf{bold}.}
\label{tab:ablation}
\begin{tabular}{lcccc}
\toprule
Benchmark & Base & \makecell{Feat.\\distill} & \makecell{Geo.\\distill} & Both \\
\midrule
VSI-Bench       & 57.64 & 60.17 & 60.42 & \textbf{62.15} \\
ScanQA (CIDEr)  & 84.61 & 89.41 & 93.13 & \textbf{93.15} \\
SQA3D           & 54.53 & 54.93 & \textbf{56.58} & 56.44 \\
SPAR-Bench      & 39.02 & 38.58 & 39.40 & \textbf{39.67} \\
SPBench-SI      & 57.07 & \textbf{62.11} & 59.00 & 59.95 \\
SPBench-MV      & 56.58 & \textbf{67.34} & 61.35 & 65.08 \\
ViewSpatial     & 50.04 & \textbf{52.68} & 49.70 & 50.60 \\
CV-Bench-2D     & 74.34 & 76.08 & 77.26 & \textbf{78.93} \\
CV-Bench-3D     & 80.88 & 79.55 & 81.14 & \textbf{83.06} \\
\midrule
Avg.\ (8 acc.)  & 58.76 & 61.43 & 60.61 & \textbf{61.98} \\
\bottomrule
\end{tabular}
\end{minipage}
\end{table*}

\subsection{Experimental Setup}
\label{sec:exp_setup}

\noindent\textbf{Implementation details.}
All variants are fine-tuned from the same base MLLM within the LLaMA-Factory
framework under the identical configuration summarized in Table~\ref{tab:impl}. The
important point for the comparisons is that every run starts from the same checkpoint,
uses the same data, optimizer, schedule, visual-token budget, and tap layers, and
changes only the transferred representation or its transfer mechanism. For explicit
geometric outputs, the pretrained DPT-style decoding head produces depth and point maps
from the tapped features.

\noindent\textbf{Training data.}
Every variant shares the same corpus of approximately $852{,}000$ examples: about
$800{,}000$ image- and video-based spatial instruction samples drawn from a mixture
of OneThinker~\cite{onethinker}, SenseNova~\cite{sensenova_si}, and
VSI-590K~\cite{cambrian_s}, together with $\approx$$52{,}000$ indoor 3D
question--answering samples from ScanQA~\cite{scanqa} and SQA3D~\cite{sqa3d}.

\noindent\textbf{Evaluation benchmarks and metrics.}
We evaluate on eight benchmarks, spanning video spatial intelligence
(VSI-Bench~\cite{vsi_bench}), indoor 3D scene QA (ScanQA~\cite{scanqa},
SQA3D~\cite{sqa3d}), image and multi-view spatial reasoning (SPAR-Bench~\cite{spar},
ViewSpatial-Bench~\cite{viewspatial}, CV-Bench~\cite{cv_bench},
SPBench~\cite{spatialladder}), and dynamic 4D reasoning (DSR-Bench~\cite{dsr_suite}).
All evaluation uses pure RGB video or images with no depth or camera pose at inference
(Section~\ref{sec:method_protocol}), and we report each benchmark's standard metric.

\begin{table*}[!tb]
\centering
\footnotesize
\setlength{\tabcolsep}{5.5pt}
\renewcommand{\arraystretch}{1.0}
\caption{VSI-Bench per-category accuracy (\%). All methods use 2D video input.
Categories are object counting (Count), absolute distance (Abs.Dist), object size
(Obj.Size), room size (Room), relative distance (Rel.Dist), relative direction
(Rel.Dir), route planning (Route), and appearance order (Appr.Order). Best among
reported entries in \textbf{bold}. $^\dagger$Numbers are reported by
GUIDE~\cite{guide}; other baseline category scores are from the VSI-Bench~\cite{vsi_bench},
VLM-3R~\cite{vlm3r}, and 3DRS~\cite{3drs} papers, or the methods' own papers.}
\label{tab:vsi}
\begin{tabular}{lccccccccc}
\toprule
Method & Avg.\ & Count & \makecell{Abs.\\Dist} & \makecell{Obj.\\Size} & \makecell{Room\\Size} & \makecell{Rel.\\Dist} & \makecell{Rel.\\Dir} & Route & \makecell{Appr.\\Order} \\
\midrule
\multicolumn{10}{l}{\textit{Proprietary}}\\
GPT-4o~\cite{gpt-4}                         & 34.0 & 46.2 & 5.3  & 43.8 & 38.2 & 37.0 & 41.3 & 31.5 & 28.5 \\
Gemini-1.5 Flash                          & 42.1 & 49.8 & 30.8 & 53.5 & 54.4 & 37.7 & 41.0 & 31.5 & 37.8 \\
Gemini-1.5 Pro                            & 45.4 & 56.2 & 30.9 & 64.1 & 43.6 & 51.3 & 46.3 & 36.0 & 34.6 \\
Gemini-2.5 Pro$^\dagger$                  & 53.6 & 46.0 & 37.4 & 68.7 & 54.4 & 62.0 & 43.9 & 47.4 & 68.8 \\
GPT-5$^\dagger$                            & 55.0 & 53.3 & 34.5 & 73.3 & 47.5 & 63.7 & 48.7 & \textbf{50.3} & 68.9 \\
\midrule
\multicolumn{10}{l}{\textit{Open-source generalist MLLMs}}\\
LongVA-7B                                 & 29.2 & 38.0 & 16.6 & 38.9 & 22.2 & 33.1 & 43.3 & 25.4 & 15.7 \\
VILA-1.5-8B                               & 28.9 & 17.4 & 21.8 & 50.3 & 18.8 & 32.1 & 34.8 & 31.0 & 24.8 \\
VILA-1.5-40B                              & 31.2 & 22.4 & 24.8 & 48.7 & 22.7 & 40.5 & 25.7 & 31.5 & 32.9 \\
LongVILA-8B                               & 21.6 & 29.1 & 9.1  & 16.7 & 0.0  & 29.6 & 30.7 & 32.5 & 25.5 \\
Qwen2.5-VL-7B~\cite{qwen25vl}             & 29.2 & 25.2 & 10.9 & 35.8 & 29.2 & 38.7 & 37.5 & 29.4 & 26.7 \\
InternVL2-2B~\cite{internvl}              & 26.5 & 25.7 & 24.0 & 20.0 & 29.2 & 32.1 & 44.1 & 30.4 & 6.3  \\
InternVL2-8B~\cite{internvl}              & 37.5 & 31.3 & 29.0 & 48.9 & 44.2 & 38.0 & 33.4 & 28.9 & 46.4 \\
InternVL2-40B~\cite{internvl}             & 37.0 & 41.3 & 26.2 & 48.2 & 27.5 & 47.6 & 32.7 & 27.8 & 44.7 \\
LLaVA-OneVision-7B~\cite{llava-onevision} & 32.4 & 47.7 & 20.2 & 47.4 & 12.3 & 42.5 & 35.2 & 29.4 & 24.4 \\
LLaVA-OneVision-72B~\cite{llava-onevision}& 40.2 & 43.5 & 23.9 & 57.6 & 37.5 & 42.5 & 39.9 & 32.5 & 44.6 \\
LLaVA-Video-7B~\cite{llavavideo}          & 35.6 & 48.5 & 14.0 & 47.8 & 24.2 & 43.5 & 42.4 & 34.0 & 30.6 \\
LLaVA-Video-72B~\cite{llavavideo}         & 40.9 & 48.9 & 22.8 & 57.4 & 35.3 & 42.4 & 36.7 & 35.0 & 48.6 \\
\midrule
\multicolumn{10}{l}{\textit{Spatial MLLMs}}\\
Spatial-MLLM~\cite{spatial_mllm}          & 47.0 & 65.3 & 34.8 & 63.1 & 45.1 & 41.3 & 46.2 & 33.5 & 46.3 \\
VG-LLM~\cite{vg_llm}                      & 47.3 & 66.0 & 37.8 & 55.2 & 59.2 & 44.6 & 45.6 & 33.5 & 36.4 \\
ViLaSR~\cite{vilasr}                      & 45.4 & 63.5 & 34.4 & 60.6 & 30.9 & 48.9 & 45.2 & 30.4 & 49.2 \\
SpatialLadder-3B~\cite{spatialladder}    & 45.7 & 63.5 & 34.3 & 61.7 & 43.9 & 45.4 & 44.8 & 35.6 & 36.4 \\
VLM-3R~\cite{vlm3r}                       & 60.9 & 70.2 & 49.4 & 69.2 & 67.1 & \textbf{65.4} & 80.5 & 45.4 & 40.1 \\
VST-7B-SFT$^\dagger$                      & 60.6 & 72.0 & 44.4 & \textbf{74.3} & \textbf{68.3} & 59.7 & 55.8 & 44.9 & 65.2 \\
Cambrian-S$^\dagger$~\cite{cambrian_s}     & 57.3 & 70.7 & 40.6 & 68.0 & 46.3 & 64.8 & 61.9 & 27.3 & \textbf{78.8} \\
3DRS~\cite{3drs}                          & 45.9 & 68.7 & 34.8 & 53.6 & 56.6 & 40.9 & 43.2 & 30.4 & 39.2 \\
\rowcolor{myblue}
\textbf{Ours}                             & \textbf{62.2} & \textbf{75.4} & \textbf{51.6} & 67.0 & 54.7 & 64.7 & \textbf{82.1} & 39.2 & 62.5 \\
\bottomrule
\end{tabular}
\end{table*}

\subsection{Ablation: Contribution of Each Transfer Component}
\label{sec:exp_ablation}

Holding the base model, data, and recipe fixed, Table~\ref{tab:ablation} isolates the
two transfer components. Each helps on its own, and each helps in a different place:
backbone-feature distillation (``Feat. distill'') is strongest on the multi-view and
single-image SPBench splits and on ViewSpatial---benchmarks that stress relational,
viewpoint-dependent cues that a free-form feature carries well---whereas
geometric-output distillation (``Geo. distill'') contributes most where metric structure
dominates, such as SQA3D. This mirrors the form-specialization finding of
Section~\ref{sec:method_typespec}.

Combining the two (``Both'') gives the best average, but the more informative fact is
where the single components fail. Feature distillation alone falls below the base model
on SPAR-Bench and CV-Bench-3D; geometric distillation alone falls below it on
ViewSpatial. The combination improves on the base in every column. Each single form has
a benchmark it costs something on, and the other form covers it---which is what it means
for the two to be more than substitutes. We therefore report ``Both'' throughout
the comparison tables, so that every benchmark is evaluated with one model rather than a
component chosen per benchmark.

The pattern also matches the diagnostics of Section~\ref{sec:method_diagnostics}. The
largest gain from feature distillation is on the multi-view split of SPBench ($+10.8$),
the setting that most directly rewards representing the same 3D point consistently across
viewpoints---the very quantity the feature diagnostic measures. The largest gain from
geometric distillation is on ScanQA ($+8.5$ CIDEr), where answers depend on metric scene
structure, which is what the geometry diagnostic measures. The two diagnostics therefore
do more than rank candidate models: each one lines up with the benchmarks its form moves
most.

\subsection{Main Results and Comparisons}
\label{sec:exp_main}

The controlled comparisons that constitute our main analysis---across the choice of
pretrained model, the representation form, and the transfer mechanism, together with the
diagnostics that explain them---are reported inline with their evidence in
Section~\ref{sec:method} (Figs.~\ref{fig:crossview}--\ref{fig:geo_mech}).

\noindent\textbf{Visual-spatial intelligence (VSI-Bench).}
On our primary benchmark (Table~\ref{tab:vsi}), the model reaches $62.2\%$ overall,
ahead of recent spatial MLLMs such as VLM-3R ($60.9$), VG-LLM ($47.3$), and
Spatial-MLLM ($47.0$), as well as far larger general models such as Gemini-2.5 Pro
($53.6$) and LLaVA-Video-72B ($40.9$), using only RGB video. All scores follow the
benchmark's own protocol, which averages the eight reporting categories with equal
weight. Per category, the model leads all reported methods on three of them, and they
divide along the two transferred forms. Counting ($75.4$) and absolute distance ($51.6$)
are single-object metric judgments, precisely what distilled depth and point maps supply.
Relative direction ($82.1$) requires relating objects seen from different viewpoints,
precisely what cross-view-consistent features supply. The design of
Section~\ref{sec:method_design} is thus visible in the category breakdown, not only in
the average.

\begin{table*}[!tb]
\centering
\scriptsize
\setlength{\tabcolsep}{1.2mm}
\renewcommand{\arraystretch}{1.02}
\caption{Dynamic 4D spatial reasoning on DSR-Bench. Baselines are from the DSR-Suite
paper. Following that setting, our entry is further fine-tuned on the DSR-Bench training
split; this is the only benchmark-specific fine-tuning setting in our evaluation. Avg.\
is sample-weighted accuracy (\%). Best in \textbf{bold}.}
\label{tab:dsr}
\resizebox{0.94\linewidth}{!}{
\begin{tabular}{lcccccccccccccc}
\toprule
\multirow{2}{*}{Method}
& \multicolumn{6}{c}{Absolute Viewpoint}
& \multicolumn{6}{c}{Relative Viewpoint}
& \multirow{2}{*}{\makecell{Non-\\Temp.}}
& \multirow{2}{*}{Avg.} \\
\cmidrule(lr){2-7} \cmidrule(lr){8-13}
& Dis. & Dir. & Ori. & Spd. & \makecell{Spd.\\Comp.} & \makecell{Dir.\\Pred.}
& Dis. & Dir. & Ori. & Spd. & \makecell{Spd.\\Comp.} & \makecell{Dir.\\Pred.}
& & \\
\midrule
\multicolumn{15}{l}{\textit{\textbf{Proprietary Models}}} \\
GPT-4o~\cite{gpt-4} & 18.8 & 29.2 & 26.8 & 29.7 & 21.5 & 26.2 & 24.1 & 23.8 & 17.2 & 32.3 & 22.8 & 24.4 & 34.7 & 26.4 \\
GPT-5 & 21.1 & 41.5 & 48.7 & 34.5 & 33.3 & 34.7 & 17.2 & 44.3 & 41.9 & 21.2 & 25.0 & 30.9 & 26.7 & 30.8 \\
Gemini-2.5-Flash & 18.8 & 27.6 & 19.5 & 25.0 & 23.6 & 22.0 & 11.2 & 28.4 & 30.8 & 23.2 & 17.1 & 22.6 & 38.8 & 24.9 \\
Gemini-2.5-Pro & 20.0 & 44.6 & 53.6 & 27.3 & 38.7 & 30.5 & 23.2 & 32.9 & 43.2 & 17.1 & 28.5 & 27.9 & 34.3 & 31.7 \\
\midrule
\multicolumn{15}{l}{\textit{\textbf{Video Understanding Models}}} \\
LLaVA-Video-7B~\cite{llavavideo} & 22.3 & 16.9 & 25.6 & 33.3 & 45.1 & 24.5 & 24.1 & 15.9 & 17.2 & 19.1 & 24.2 & 21.4 & 33.9 & 25.9 \\
VideoRefer & 23.5 & 18.4 & 25.6 & 33.5 & 45.4 & 27.1 & 25.0 & 16.1 & 18.5 & 20.2 & 26.4 & 22.6 & 34.7 & 26.9 \\
LongVILA-R1 & 20.0 & 21.5 & 23.1 & 21.4 & 37.6 & 22.8 & 24.1 & 26.1 & 28.3 & 22.2 & 17.8 & 20.8 & 33.9 & 25.3 \\
\midrule
\multicolumn{15}{l}{\textit{\textbf{General-Purpose Models}}} \\
Qwen2.5-VL-7B~\cite{qwen25vl} & 18.8 & 15.3 & 14.6 & 42.8 & 29.0 & 19.4 & 31.8 & 19.3 & 11.1 & 22.2 & 19.2 & 20.2 & 30.1 & 23.5 \\
Qwen2.5-VL-32B~\cite{qwen25vl} & 31.7 & 21.5 & 23.1 & 44.0 & 36.5 & 25.4 & 27.5 & 23.8 & 37.0 & 27.2 & 29.2 & 21.4 & 36.2 & 29.9 \\
Qwen3-VL-8B-Instruct & 23.5 & 24.6 & 42.6 & 29.7 & 27.9 & 33.8 & 18.1 & 28.4 & 34.5 & 24.2 & 22.1 & 27.9 & 33.5 & 28.7 \\
Qwen3-VL-30B-A3B-Instruct & 25.8 & 27.6 & 46.3 & 30.9 & 31.1 & 34.7 & 20.6 & 29.5 & 37.0 & 28.2 & 24.2 & 31.5 & 35.4 & 31.1 \\
InternVL3.5-8B~\cite{internvl} & 23.5 & 27.6 & 28.0 & 34.5 & 24.7 & 27.9 & 22.4 & 17.0 & 19.7 & 28.2 & 30.0 & 14.2 & 30.1 & 25.4 \\
InternVL3.5-38B~\cite{internvl} & 25.8 & 27.8 & 29.2 & 34.2 & 24.7 & 28.5 & 26.7 & 16.3 & 23.4 & 29.2 & 32.1 & 15.4 & 31.3 & 26.7 \\
\midrule
\multicolumn{15}{l}{\textit{\textbf{Spatial Reasoning Models}}} \\
VG-LLM~\cite{vg_llm} & 55.2 & 32.3 & 58.5 & 57.1 & 51.6 & 32.2 & 56.0 & 36.3 & 32.0 & 30.3 & 32.1 & 29.1 & 27.9 & 38.4 \\
DSR-Suite~\cite{dsr_suite} & 87.0 & 73.8 & 84.1 & 73.8 & 72.0 & 35.5 & 75.8 & 76.1 & 77.7 & 60.6 & 37.1 & 35.1 & 46.4 & 58.9 \\
\rowcolor{myblue}
\textbf{Ours} & \textbf{91.8} & \textbf{92.3} & \textbf{92.7} & \textbf{81.0} & \textbf{86.0} & \textbf{46.2} & \textbf{86.6} & \textbf{93.2} & \textbf{92.5} & \textbf{86.7} & \textbf{57.1} & \textbf{43.1} & \textbf{49.8} & \textbf{70.7} \\
\bottomrule
\end{tabular}
}
\end{table*}

\begin{table}[!tb]
\centering
\footnotesize
\setlength{\tabcolsep}{5pt}
\renewcommand{\arraystretch}{1.05}
\caption{Indoor 3D scene question answering on ScanQA val and SQA3D test. ScanQA
reports CIDEr (C) and exact match (EM); SQA3D reports accuracy (\%). Input indicates
explicit 3D/posed RGB-D, reconstructed BEV from 3D geometry, or RGB video. Best per
column in \textbf{bold}; ``--'' denotes not reported.}
\label{tab:sceneqa}
\begin{tabular}{llccc}
\toprule
Method & Input & C & EM & SQA3D \\
\midrule
ScanQA~\cite{scanqa}                  & 3D & 64.9  & 21.1 & 47.2 \\
3D-LLM (FlanT5)~\cite{3d_llm}         & 3D & 69.4  & 20.5 & --   \\
3D-VisTA~\cite{3dvista}               & 3D & 69.6  & 22.4 & 48.5 \\
Grounded 3D-LLM~\cite{grounded3dllm}  & 3D & 75.9  & --   & --   \\
LL3DA~\cite{ll3da}                    & 3D & 76.8  & --   & --   \\
Scene-LLM~\cite{scenellm}             & 3D & 80.0  & 27.2 & 54.2 \\
Chat-Scene~\cite{chatscene}           & 3D & 87.7  & 21.6 & 54.6 \\
LLaVA-3D~\cite{llava3d}               & 3D & 91.7  & 27.0 & 55.6 \\
LEO-VL~\cite{leovl}                   & 3D & 101.1 & --   & 58.7 \\
LEO~\cite{leo}                        & 3D & 101.4 & 24.5 & 50.0 \\
Video-3D LLM~\cite{video3dllm}        & 3D & \textbf{102.1} & \textbf{30.1} & 58.6 \\
GPT4Scene~\cite{gpt4scene}            & 3D-BEV & 96.3  & 28.2 & \textbf{59.4} \\
\midrule
Qwen2-VL-7B~\cite{qwen2vl}            & 2D & 53.9  & 19.0 & 40.7 \\
Oryx-34B~\cite{oryx}                  & 2D & 72.3  & --   & 50.9 \\
LLaVA-Video-7B~\cite{llavavideo}      & 2D & 88.7  & --   & 48.5 \\
Spatial-MLLM~\cite{spatial_mllm}      & 2D & 91.8  & 26.3 & 55.9 \\
\rowcolor{myblue}
\textbf{Ours} & 2D & 93.2 & 26.6 & 56.4 \\
\bottomrule
\end{tabular}
\end{table}

\begin{table}[!tb]
\centering
\scriptsize
\setlength{\tabcolsep}{2.6pt}
\renewcommand{\arraystretch}{1.1}
\caption{Image and multi-view spatial reasoning (2D input). SPAR/ViewSpatial: overall
accuracy; CV-Bench: 2D/3D splits; SPBench: single-image (SI)/multi-view (MV) splits
(all \%). SPAR uses the full benchmark where reported; Claude-3.7-Sonnet is the
tiny/API result reported by SPAR because the full result is unavailable. Best per column
in \textbf{bold}; ``--'' not reported.}
\label{tab:spatial}
\begin{tabular}{lcccccc}
\toprule
Method & SPAR & \makecell{View-\\Spatial} & \multicolumn{2}{c}{CV-Bench} & \multicolumn{2}{c}{SPBench} \\
\cmidrule(lr){2-2}\cmidrule(lr){3-3}\cmidrule(lr){4-5}\cmidrule(lr){6-7}
 & Avg & Overall & 2D & 3D & SI & MV \\
\midrule
LLaVA-OneVision-7B~\cite{llava-onevision} & 31.2 & 27.5 & 53.2 & 63.5 & 33.2 & 32.2 \\
Llama-3.2-Vision-11B~\cite{viewspatial}   & -- & 28.8 & -- & -- & -- & -- \\
LLaVA-NeXT-Video-7B~\cite{viewspatial}    & -- & 30.6 & -- & -- & -- & -- \\
GPT-4o~\cite{gpt-4}                        & 38.1 & 35.0 & 69.4 & 81.3 & 42.4 & 48.2 \\
Claude-3.7-Sonnet~\cite{spar}             & 21.8 & -- & -- & -- & -- & -- \\
Gemini-2.0 Flash                          & -- & 32.6 & -- & -- & 54.7 & 51.4 \\
Qwen2.5-VL-3B~\cite{qwen25vl}             & 24.6 & 35.9 & 69.1 & 72.2 & 40.3 & 36.6 \\
Qwen2.5-VL-7B~\cite{qwen25vl}             & 33.1 & 36.9 & 75.0 & \textbf{83.1} & 48.4 & 37.3 \\
Qwen2.5-VL-72B~\cite{qwen25vl}            & 37.0 & -- & -- & -- & -- & -- \\
InternVL2.5-4B~\cite{internvl}            & 30.6 & 37.9 & 73.5 & 75.1 & 42.5 & 43.2 \\
InternVL2.5-8B~\cite{internvl}            & 36.3 & 43.2 & -- & -- & 42.3 & 41.4 \\
InternVL3-14B~\cite{internvl}             & -- & 40.3 & -- & -- & -- & -- \\
Kimi-VL-A3B~\cite{spatialladder}          & 29.4 & 33.6 & 41.9 & 54.7 & 35.3 & 37.0 \\
Cambrian-8B~\cite{cv_bench}               & -- & -- & 72.3 & 72.0 & -- & -- \\
Video-R1~\cite{video-r1}                  & -- & -- & -- & -- & 44.8 & 40.7 \\
Spatial-MLLM~\cite{spatial_mllm}          & -- & -- & -- & -- & 43.7 & 61.8 \\
ViLaSR~\cite{vilasr}                      & 37.6 & -- & -- & -- & 50.2 & 61.8 \\
SpaceR~\cite{spacer}                      & 37.6 & -- & -- & -- & 48.6 & 59.4 \\
\midrule
\rowcolor{myblue}
\textbf{Ours} & \textbf{39.7} & \textbf{50.6} & \textbf{78.9} & \textbf{83.1} & \textbf{60.0} & \textbf{65.1} \\
\bottomrule
\end{tabular}
\end{table}

\noindent\textbf{Dynamic 4D reasoning (DSR-Bench).}
DSR-Bench targets dynamic, 4D spatial reasoning, a setting distinct from the static
benchmarks above. Following the DSR-Suite protocol, we take our spatially-trained
model and \emph{further fine-tune it on the DSR-Bench training split} before
evaluation; the remaining rows are as reported in the DSR-Suite paper, where the general
and proprietary models are evaluated zero-shot. DSR-Suite is therefore the directly
comparable entry, being trained on the same split. Across the $13$ subtasks under
absolute and relative
viewpoints (Table~\ref{tab:dsr}), our model reaches $70.7\%$ overall, $+11.8$ over the
official DSR-Suite ($58.9$) and far above the zero-shot general and proprietary models
(e.g.,~Gemini-2.5-Pro $31.7$, GPT-5 $30.8$, VG-LLM $38.4$). The gains hold across
\emph{every} subtask and are largest on the dynamic ones---relative speed ($+26.1$),
relative speed comparison ($+20.0$), absolute direction ($+18.5$), and relative
direction ($+17.1$)---showing that a model built by transferring 3D structure carries
that advantage into reasoning about motion and changing viewpoints, not only static
scenes.

\noindent\textbf{Indoor 3D scene QA (ScanQA, SQA3D).}
Although these benchmarks are usually addressed with explicit 3D input, our pure-RGB
model is highly competitive with both 3D-input and 2D methods
(Table~\ref{tab:sceneqa}). It reaches $93.2$ CIDEr on ScanQA and $56.4\%$ on SQA3D.
On ScanQA, it exceeds older 3D-input LLMs such as 3D-LLM, LL3DA, Grounded 3D-LLM,
Scene-LLM, Chat-Scene, and LLaVA-3D; on SQA3D, it is above the reported results of
Scene-LLM, Chat-Scene, and LLaVA-3D. Methods built specifically for 3D scene QA---via
explicit 3D input, object-centric scene tokens, or reconstruction supervision---can
report higher scores (e.g.,~Video-3D LLM, LEO-VL, VLM-3R, and recent
reconstruction-tuned models~\cite{vidllm,reg3d}), but they rely on geometry or
reconstruction that our pure-RGB setting does not use. Three regimes are worth keeping
apart here: methods given external 3D at inference, methods that infer geometry from RGB
at inference, and methods for which geometry appears only as training supervision. Ours
is the last; the entries above belong to the first two, so this table positions the
finished system against the literature rather than isolating the effect of transfer,
which Table~\ref{tab:ablation} measures directly. Among
general 2D video models it leads clearly (e.g.,~LLaVA-Video-7B $88.7$ / $48.5$,
Oryx-34B $72.3$ / $50.9$), and it is slightly above
the video-input Spatial-MLLM ($91.8$ / $55.9$). This supports the paper's claim that an
RGB model carrying enough 3D structure can rival methods given geometry directly.

\begin{figure*}[!t]
\centering
\figtrim{0.84\textwidth}{46.03}{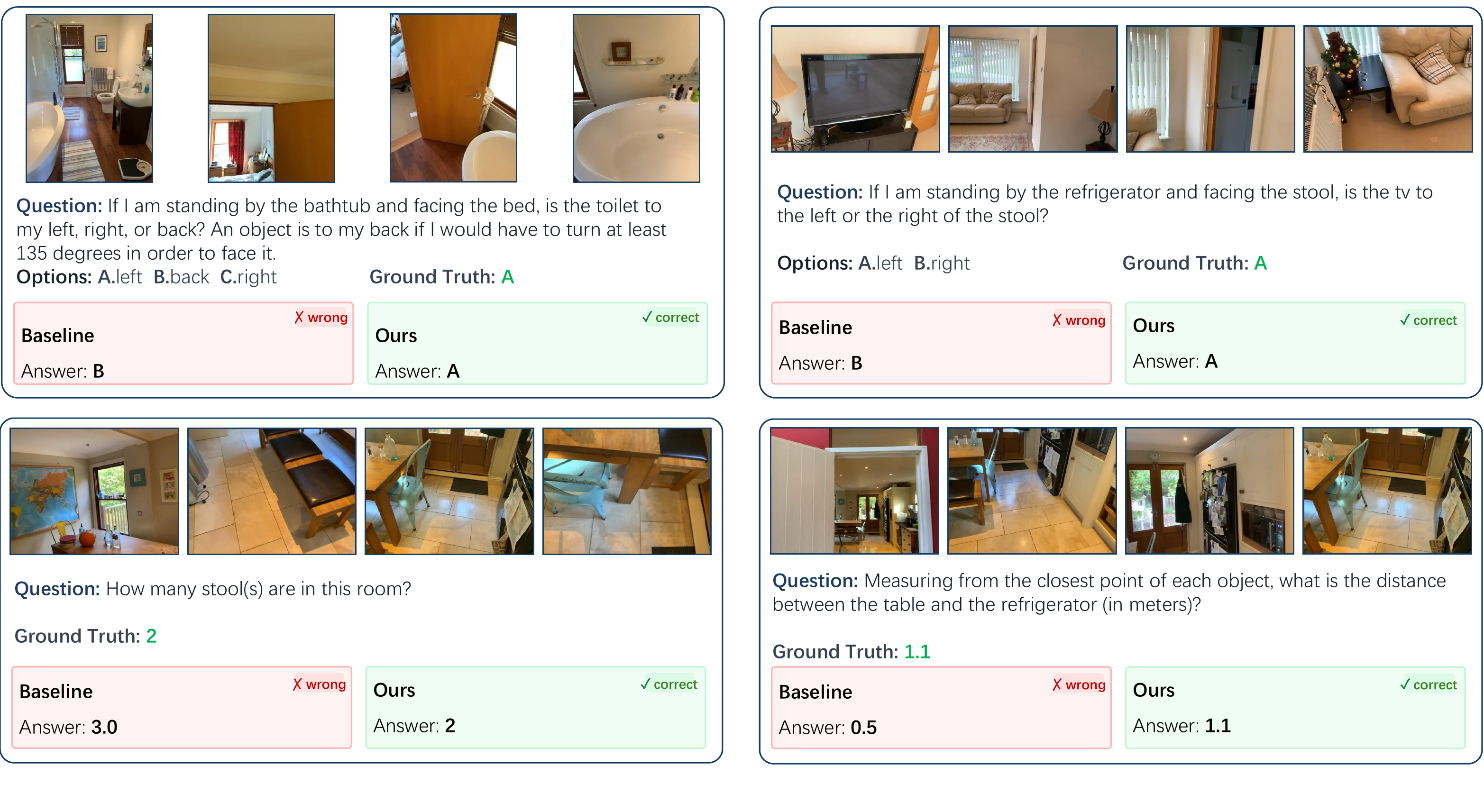}
\caption{\textbf{Qualitative spatial QA examples.} Each panel shows input views, the
question, ground truth, and answers from the base model and our model. The examples
show cases where the base model's answer is inconsistent with the depicted scene layout,
whereas our transferred-representation model recovers the annotated answer under the same
RGB input.}
\label{fig:qual_qa}
\end{figure*}

\begin{figure*}[!t]
\centering
\figtrim{0.8\textwidth}{1.52}{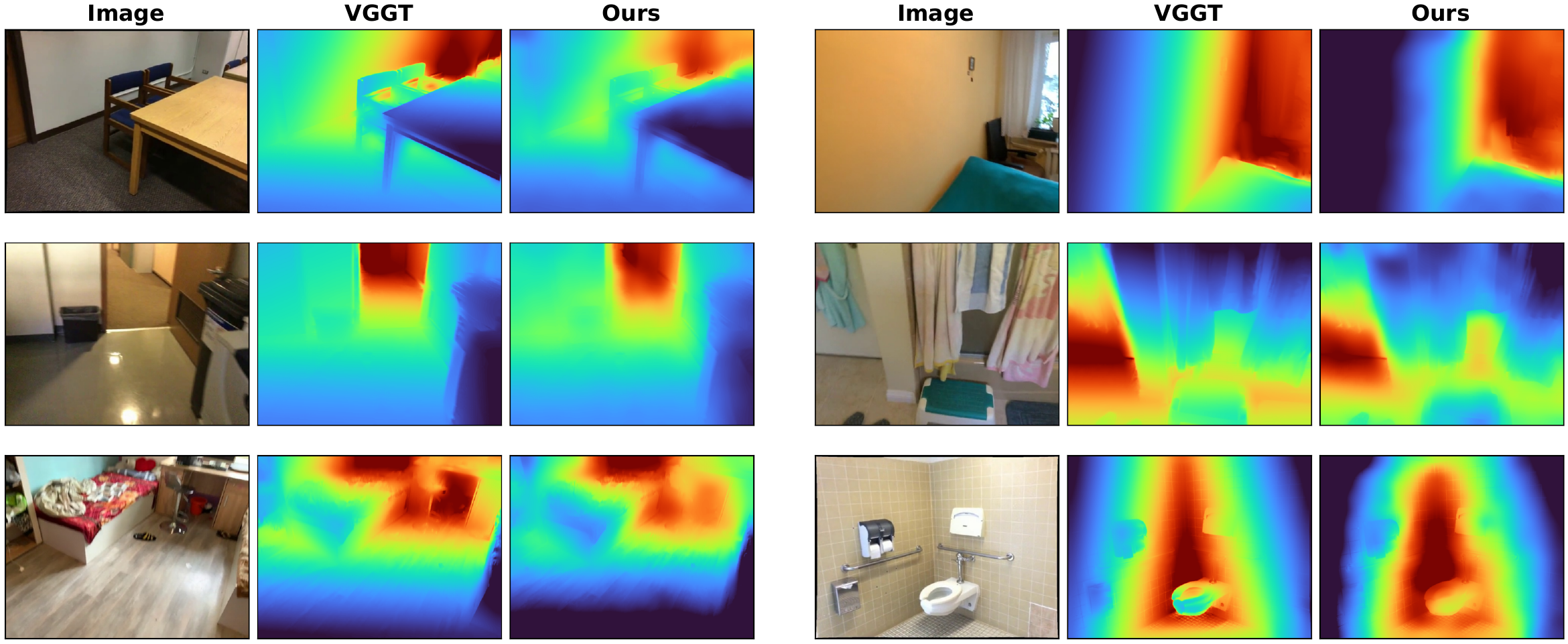}
\caption{\textbf{Qualitative depth estimation.} Each triplet: input image, VGGT~\cite{vggt} depth,
and our DPT-decoded depth. Our output has the same geometric form as VGGT's depth
prediction, but the model is trained primarily to improve spatial understanding rather
than to serve as a standalone geometry predictor; the decoded depth is therefore
smoother, while still preserving plausible depth ordering.}
\label{fig:qual_depth}
\end{figure*}

\begin{figure*}[!t]
\centering
\figtrim{0.8\textwidth}{1.52}{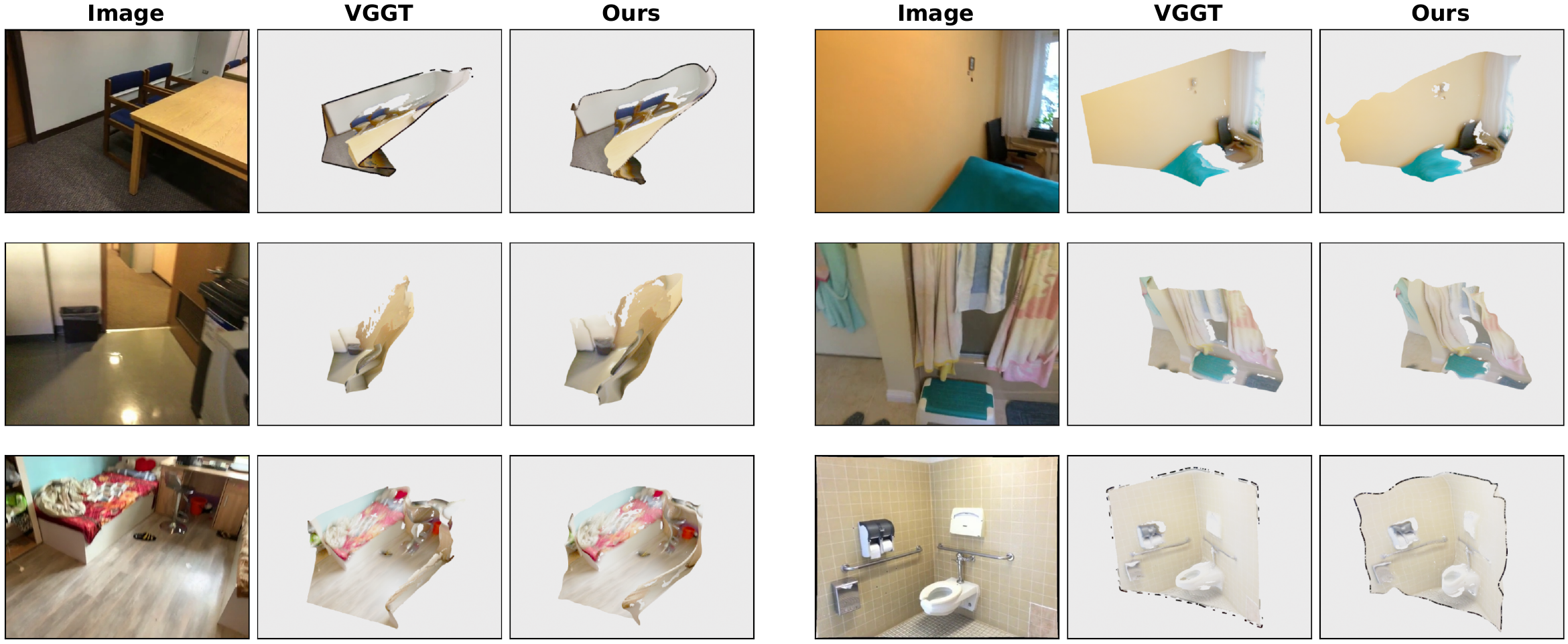}
\caption{\textbf{Qualitative 3D reconstruction (point maps)}. Layout as in Fig.~\ref{fig:qual_depth}:
input image, VGGT~\cite{vggt} reconstruction, and our DPT-decoded point map. The point
map is analogous to VGGT's geometric output but is produced by a spatial-understanding
model rather than a dedicated reconstruction model, so it is coarser while still
recovering coherent room-level structure.}
\label{fig:qual_recon}
\end{figure*}

\noindent\textbf{Image and multi-view spatial reasoning.}
Across the four image/multi-view benchmarks in Table~\ref{tab:spatial}, our model is best
or tied-best on every column. On SPAR-Bench it reaches $39.7$,
above GPT-4o ($38.1$), Qwen2.5-VL-72B ($37.0$), and all $7$B/$8$B baselines; on
ViewSpatial-Bench it scores
$50.6$, well above the best general model InternVL2.5-8B ($43.2$); on CV-Bench it is
best on the 2D split ($78.9$) and ties the best on the 3D split ($83.1$); and on
SPBench it attains $60.0$/$65.1$ (SI/MV), above the general models and recent 7B
spatial baselines such as SpaceR ($48.6$/$59.4$), ViLaSR ($50.2$/$61.8$), and
Spatial-MLLM ($43.7$/$61.8$).

\subsection{Qualitative Analysis}
\label{sec:exp_qual}

\noindent\textbf{Spatial QA examples.}
Fig.~\ref{fig:qual_qa} complements the aggregate benchmark results with representative
held-out QA cases. The comparison is between the base model and our final
transferred-representation model. In these
examples, the relevant scene content is visible in the RGB views, but the answer requires
organizing that content into a 3D layout. The base model's answers are inconsistent with
the layout the views depict, whereas the transferred model recovers the annotated relation
from the same frames.

\noindent\textbf{Decoded geometric outputs.}
Because our geometric variant is trained to decode explicit depth and point maps through
the DPT-style decoding head (Section~\ref{sec:method}), we can inspect the geometry it
learns directly. Figs.~\ref{fig:qual_depth} and~\ref{fig:qual_recon} compare these
decoded outputs against VGGT~\cite{vggt} on held-out indoor scenes, illustrating the
geometry that the transferred representation makes available to the MLLM.

\section{Conclusion}
\label{sec:conclusion}

We have studied which visual representations serve spatial understanding in multimodal
large language models, and why. Holding the base model, data, and protocol fixed, we
varied where a representation comes from, what form it takes, and how it is transferred,
and asked what property of a representation explains the differences. Two regularities emerged. The amount of 3D structure a
representation carries correlates with the gain it delivers, measured as cross-view
consistency for latent features and as depth and reconstruction accuracy for explicit
geometry. The appropriate transfer mechanism, in turn, follows from the representation's form. Guided by these regularities, we distilled
both forms into one model, which runs on RGB video alone and improves on every benchmark
of the shared-recipe comparison.

The practical value of the first regularity is that it can be evaluated before any MLLM
is trained. Choosing a representation is otherwise an expensive search, since each
candidate costs a full fine-tuning run; a property computed from the representation alone
reduces that search to a measurement, and the same measurement anticipates which spatial
skills the resulting model will be strongest at. The second regularity is a design rule
rather than a measurement: the two forms are not interchangeable inputs to a common
interface, and explicit geometry in particular is best internalized during training
rather than handed to the language model at inference.

\noindent\textbf{Limitations and future work.} Our diagnostics are established on a moderate pool of pretrained models and on static indoor scenes; extending them to more representations and to dynamic scenes, for instance with a 4D counterpart of cross-view consistency, would test how far the regularities generalize. Our decoding-head comparison also shows that the pretrained DPT-style configuration matters without isolating whether architecture, initialization, or trainability is responsible. More broadly, as in most work on spatial representations for MLLMs, 3D structure here enters only at the fine-tuning stage. With sufficient compute, data, and high-quality 3D annotation, the same learning could instead take place during pretraining, so that a multimodal model acquires 3D perception and reasoning natively, rather than having them added afterwards.


\section*{Acknowledgment}
This work was supported by the Hong Kong Research Grants Council (General Research
Fund, Project No.~17213825), the Hong Kong Innovation and Technology Commission
(Innovation and Technology Fund, Project No.~ITS/488/24FP), and the HKU Seed Fund
for PI Research.

\bibliographystyle{plain}
\bibliography{main}

\end{document}